\documentclass[runningheads]{llncs}
\usepackage{amsmath}
\usepackage{bm}
\usepackage{algorithm}
\usepackage{algorithmic}
\usepackage{array}
\usepackage{multirow}
\usepackage[T1]{fontenc}
\usepackage{graphicx}
\usepackage[table]{xcolor}
\usepackage[most]{tcolorbox}
\tcbuselibrary{skins,breakable}
\usepackage{xparse}
\usepackage{placeins}   
\usepackage{array}      
\usepackage{enumitem}
\usepackage{adjustbox}

\definecolor{lightblue}{HTML}{E3F2FD}
\definecolor{darkblue}{HTML}{1976D2}

\NewDocumentEnvironment{promptbox}{O{} +b}{%
  \begin{tcolorbox}[
    enhanced,
    width=\linewidth,
    colback=lightblue,
    colframe=darkblue,
    boxrule=0.6pt,
    arc=2pt,
    title={#1},
    fonttitle=\bfseries,
    colbacktitle=darkblue,
    coltitle=white,
    boxed title style={sharp corners, boxrule=0pt},
    left=6pt,right=6pt,top=6pt,bottom=6pt,
    before upper=\raggedright\setlength{\parindent}{0pt}\sloppy
  ]
  #2
  \end{tcolorbox}
}{}

\usepackage{booktabs}
\usepackage{tabularx}
\usepackage{capt-of}

\begin{document}
\title{HiMeS: Hippocampus-inspired Memory System for Personalized AI Assistants}
\titlerunning{HiMeS}
%
%
\author{Hailong Li \and Feifei Li \and Wenhui Que \and Xingyu Fan}
\authorrunning{H. Li et al.}
%
\institute{WeChat, Tencent Inc., Beijing, China\\
\email{\{aloneli, niyali, victorque, fanxfan\}@tencent.com}}
\maketitle              

\begin{abstract}
Large language models (LLMs) power many interactive systems such as chatbots, customer-service agents, and personal assistants. In knowledge intensive scenarios requiring user specific personalization, conventional retrieval-augmented generation (RAG) pipelines exhibit limited memory capacity and insufficient coordination between retrieval mechanisms and user specific conversational history, leading to redundant clarification, irrelevant documents, and degraded user experience. 
Inspired by the hippocampus–neocortex memory mechanism, we propose HiMeS, an AI-assistant architecture that fuses short-term and long-term memory. Our contributions are fourfold: (1) A short-term memory extractor is trained end-to-end with reinforcement learning to compress recent dialogue and proactively pre-retrieve documents from the knowledge base, emulating the cooperative interaction between the hippocampus and prefrontal cortex. (2) A partitioned long-term memory network stores user-specific information and re-ranks retrieved documents, simulating distributed cortical storage and memory reactivation. (3) On a real-world industrial dataset, HiMeS significantly outperforms a cascaded RAG baseline on question-answering quality. (4) Ablation studies confirm the necessity of both memory modules and suggest a practical path toward more reliable, context-aware, user-customized LLM-based assistants.


\end{abstract}

\section{Introduction}
Recent advances in large language models (LLMs) have facilitated the deployment of agent-based interactive systems in industrial scenarios such as customer service and personal assistants~\cite{Agents4MTConversations}. The combination of retrieval-augmented generation (RAG) techniques~\cite{RAG} and long-context extrapolation~\cite{longcontext1} has further expanded the capacity of LLM agents to handle knowledge-intensive tasks. As LLMs evolve, AI assistants can now generate responses on behalf of content authors.

In our industrial deployment~\cite{wechat}, diverse entities including individual users, media outlets, enterprises, and governmental agencies can register accounts on the platform, publish content, and engage with interested users.
The platform hosts an extensive repository of knowledge articles, yet when users seek information from an account's previously published content, they must manually search for and retrieve the relevant articles themselves.
This approach proves inefficient and yields suboptimal accuracy, motivating the deployment of AI assistants for accounts.

However, existing methods encounter significant challenges when directly applied to this industrial scenario.

\begin{itemize}
    \item \textbf{Incomplete exploitation of short-term memory (current dialogue).}
    Contemporary AI assistant systems typically follow a sequential pipeline: RAG retrieval based on the current query, followed by prompt assembly incorporating dialogue history and retrieval results, and finally response generation~\cite{RAG-3}. In practice, retrieval queries frequently omit critical information from preceding dialogue turns, whereas human customer service agents systematically review conversational context before formulating responses. This omission creates a semantic mismatch between the query and the user’s true information need and lowers retrieval efficiency. Consequently, we compress essential short-term context into refined queries, enabling more contextually grounded retrieval.
    
    \item \textbf{Catastrophic forgetting of long-term memory (historical dialogue).}
    LLM-based agents continue to face challenges in preserving information across multiple sessions~\cite{long-term-agents}. Repeated clarifications and frequent topic shifts lead to memory loss or confusion, resulting in redundant dialogue turns and degraded user experience. A primary factor is that interaction data are typically discarded upon session termination~\cite{forgetting-in-agents}, preventing knowledge-intensive assistants from integrating stable user profiles with current queries in the manner of human experts~\cite{user-profile}. To address this issue, we equip the assistant with persistent user profiles that integrate long- and short-term memories when filtering, retrieving, and recommending documents, thereby producing more informed and customized replies~\cite{user-profile-1,user-profile-2,user-profile-3}. In our production deployment, we quantify long-term memory degradation through the Repeated Asking Rate (RAR), defined as the proportion of questions repeated across sessions. Prior to implementing long-term memory capabilities, RAR reached 70–80\%, directly motivating the development of the long-term memory module.
\end{itemize}

To better understand these challenges, we examined how human experts use long-term user profiles and recent dialogue history when answering questions. Unlike current AI assistants, humans progressively construct internal profiles of conversation partners, relying primarily on immediate conversational cues for new users while drawing upon accumulated impressions for returning users. This behaviour parallels the hippocampus–cortex memory system~\cite{Hippocampus}, in which short-term experiences are stored and later consolidated into long-term memories that can be reactivated when needed.

Motivated by this neural mechanism, we present HiMeS, a memory framework for AI assistants that emulates the cooperation between hippocampus and cerebral cortex.
HiMeS leverages the information-compression capabilities of large language models (LLMs) to distill recent dialogue into refined retrieval queries, thereby enhancing contextual relevance~\cite{query-rewrite-2}.
A distributed storage system maintains comprehensive archives of each user's historical queries and, upon receiving new queries, retrieves relevant historical interactions to re-rank document chunks obtained through short-term memory processing, retaining only the most fine-grained and relevant knowledge~\cite{long-term-memory}.
By integrating short-term and long-term memory systems, the AI assistant achieves more reliable recall of user-specific information compared to human experts. While human long-term memory exhibits inherent imprecision, distributed storage systems preserve information with high fidelity for subsequent retrieval.(Figure~1).

\begin{figure}
\centering
\includegraphics[width=\textwidth]{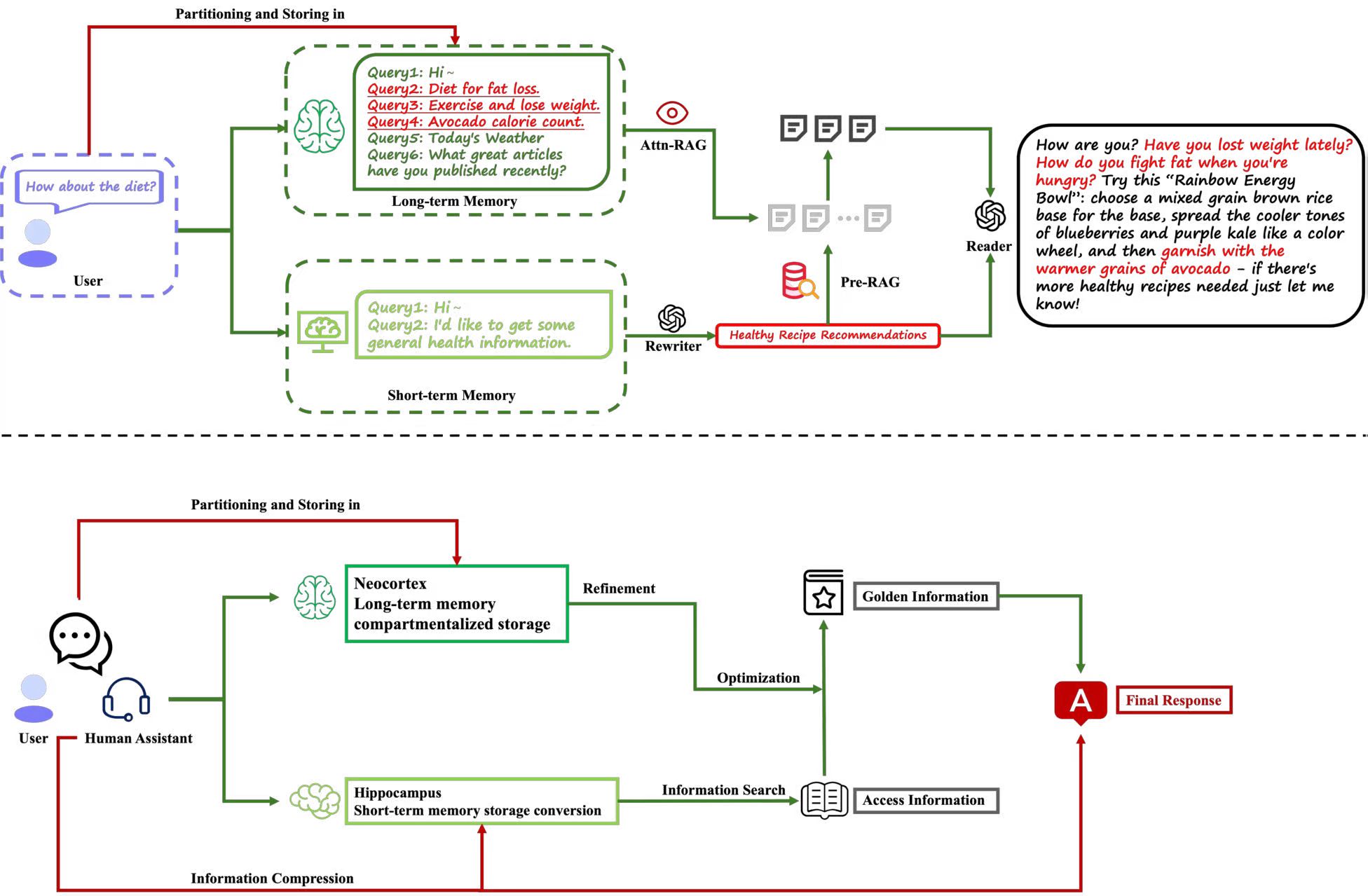} 
\caption{Illustration of HiMeS}
\label{fig1}
\end{figure}

To validate the optimization achieved by HiMeS in production interaction scenarios, we conducted extensive experiments on an industrial multi-turn dialogue benchmark and found that HiMeS significantly outperforms traditional cascaded RAG systems utilizing advanced LLMs such as DeepSeek-R1 and DeepSeek-V3 in knowledge-intensive question answering, indicating substantial potential for continued advancement.

Our main contributions are summarised as follows:
\begin{itemize}
    \item \textbf{To capture the complete conversational semantics embedded in users' short-term dialogues, we develop an end-to-end reinforcement learning-based query rewriting model trained with multi-source reward signals.} These reward components are integrated through weighted combination, with weighting coefficients determined and validated through systematic ablation studies. Given recent dialogue histories as input, the model compresses these short-term memories into long-term memory representations and generates a rewritten query, which is then used to retrieve relevant documents from the knowledge base, thereby modelling the cooperative interaction between the hippocampus and the prefrontal cortex.
    \item \textbf{To enable the AI assistant to exploit accumulated information efficiently, we build a partitioned long-term memory storage network that maps a user’s historical queries to topic-specific partitions.} When a new query arrives, the relevant memories are activated to interact with the retrieved documents and allocate differentiated attention to them, thereby modelling cortical operations that distribute long-term information across the temporal, parietal and frontal lobes.
    \item On a real-world industrial test dataset derived from our deployment scenario, HiMeS achieves substantial improvements in question-answering performance compared to cascaded RAG baselines, demonstrating the usefulness of our method in such scenarios. 
    \item Ablation studies confirm the necessity of both short-term and long-term memory modules, offering a novel framework for constructing more reliable, context-aware, and user-customized LLM-based AI assistant systems. Furthermore, we demonstrate the transferability of the memory system across different LLMs and application domains.
\end{itemize}
\section{Related Work }

\subsection{Short-term memory and dialogue compression}

Recent research has investigated methods for compressing conversational context prior to input to dialogue policies or LLMs. Early studies drew on state compression in reinforcement learning (RL): Paischer et al.~\cite{RL-history-compress-1} proposed HELM, which employs a frozen Transformer to summarize historical conversations, enabling RL policies to operate under partial observability conditions. With the rise of large language models, summarisation-based methods became dominant. Wang et al.~\cite{summarization} proposed Recursively Summarizing Enables Long-Term Dialogue Memory, which builds hierarchical summaries to keep a bounded context window in ultra-long conversations. Prompt-optimisation methods such as LLM-Lingua~\cite{LLMLingua} prune low-entropy tokens to achieve high compression ratios, while MemPrompt~\cite{MemPrompt} maintains a micro-memory updated with user feedback to correct misunderstandings in subsequent turns. While these approaches demonstrate effective compression of short-term dialogue, most treat summarization, token pruning, or prompt editing as isolated modules. Specifically, they lack end-to-end training with downstream tasks, rarely integrate knowledge-base retrieval, and consequently fail to fully leverage the short-term memory and information-compression capabilities of LLMs.

\subsection{Long-term memory and memory storage}

Retrieval-augmented language models (RAG) pioneered the exploration of exploring external long-term memory. For agent-style applications, Generative Agents ~\cite{Generative-Agents}, LongMem~\cite{LongMem}, and MemGPT~\cite{MemGPT} store experiences as vectors and activate only the most salient memories based on relevance, recency, or importance, often via multi-stage reranking or hierarchical memory management. Despite this progress, current long-term memory solutions remain largely decoupled from short-term compression modules, rely predominantly on vector similarity for relevance estimation, and have received limited systematic evaluation in production industrial systems.

\section{Method}

\subsection{Short-term memory module}

A conventional approach for endowing assistants with short-term memory compression involves training a dedicated rewriter model on multi-turn dialogue-rewriting data annotated by humans or LLMs, then integrating it prior to the retrieval stage of the assistant pipeline. As complex agent systems, LLM-based knowledge-intensive RAG question-answering assistants typically comprise multiple independently optimized modules. This practice fragments system-wide optimization objectives, preventing effective alignment with the goal of generating high-quality responses.
Consequently, we posit that optimization metrics for query-rewriting models should directly reflect downstream task performance (e.g., QA accuracy or conversational response quality) rather than surface-level rewriting quality. A frozen response model (the Task LLM) serves as the reference standard for task execution, with its outputs providing quantitative evaluation of whether rewritten prompts and queries genuinely improve end-to-end performance.
We adopt an end-to-end alignment training pipeline consisting of supervised fine-tuning (SFT) followed by reinforcement learning from human feedback (RLHF).

During the SFT stage, we align output distributions and formats using high-quality multi-turn query-rewriting data generated by a multi-agent system. At this stage, we systematically vary data sources and mixing ratios for query-rewriting samples, selecting optimal configurations based on validation performance.
In the RLHF stage, we implement an end-to-end reward modeling scheme based on Group Relative Policy Optimization (GRPO) to simulate RAG retrieval and response generation during inference, jointly modeling rewards for both query-rewriting and response generation components (Figure 2). As a lightweight PPO variant that compares samples within a group instead of learning a separate value network (Critic)~\cite{grpo-deepseek}, GRPO is well suited to our RLHF setup.
For the reward, we design a scheme called ``hard supervised explicit reward'' (HSER). HSER is computed from the black-box model’s response \( A_{\text{pred}} \) and the golden reference \( A_{\text{label}} \). We obtain the Rouge-L F1 score \( F1_{h} \), the exact-match score \( EM_{h} \), and the hit score \( Hit \), which indicates whether the model response appears in the RAG-retrieved content \( C \). These three components are then integrated to compute the overall reward value (Algorithm 1), with weights \( \alpha \) and \( \beta \) determined through comparative experiments evaluating trained model performance on the test set. 

\begin{algorithm}
\caption{Reward Modeling}
\label{alg:reward}
\small
\begin{tabular}{@{}>{\raggedright\arraybackslash}p{0.32\linewidth} p{0.64\linewidth}@{}}
\textbf{Initialize:} &
HSER components $F1_h$, $EM_h$, $Hit$.\\[0.6ex]

\textbf{Compute $F1_h$:} &
$F1_h \gets \text{Rouge-L F1}(A_{\text{pred}}, A_{\text{label}}).$\\[0.8ex]

\textbf{Compute $EM_h$:} &
$EM_h \gets
\begin{cases}
1, & \text{if } A_{\text{pred}} \equiv A_{\text{label}},\\
0, & \text{otherwise}.
\end{cases}$\\[1.2ex]

\textbf{Compute $Hit$:} &
$Hit \gets
\begin{cases}
1, & \text{if } A_{\text{pred}} \in \bm{C} \text{ (RAG-retrived contents)},\\
0, & \text{otherwise}.
\end{cases}$\\[1.2ex]

\textbf{Calculate total reward:} &
$\mathrm{reward} \gets F1_h + \alpha\, EM_h + \beta\, Hit.$\\
\end{tabular}
\normalsize
\end{algorithm}

\begin{figure}
\centering
\includegraphics[width=\textwidth]{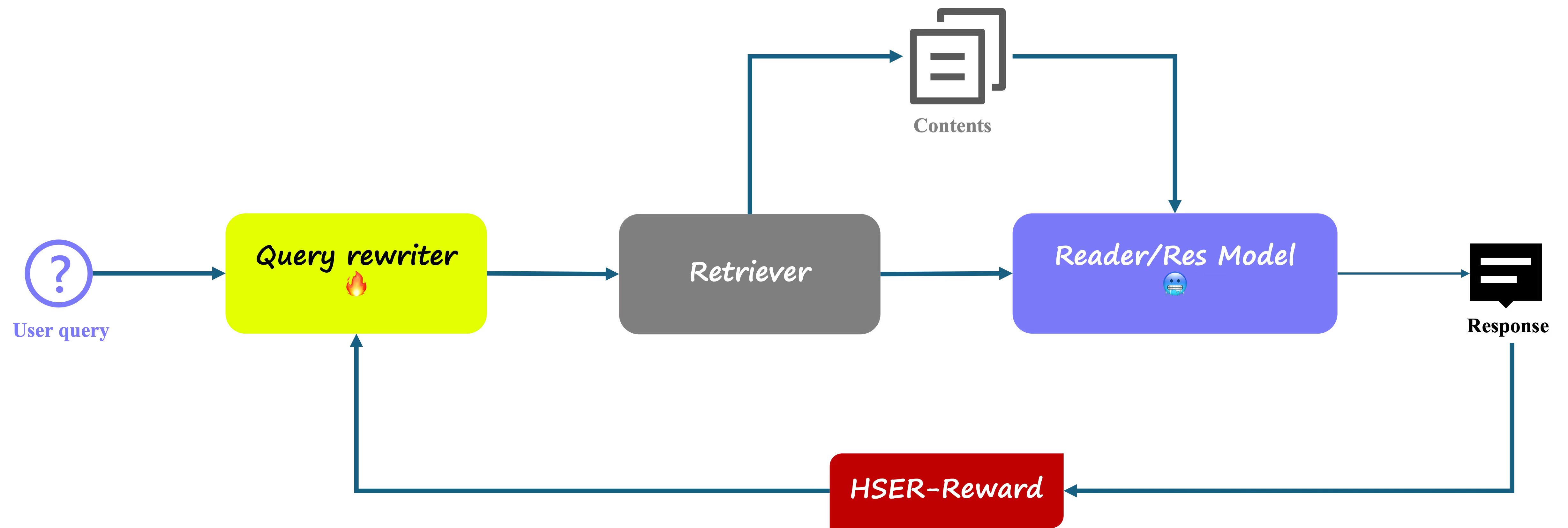} 
\caption{RLHF Training Pipeline of Query Rewriter}
\label{fig2}
\end{figure}

Through targeted fine-tuning for dialogue-history compression, our assistant system acquires short-term memory capabilities, enabling it to review conversational history analogously to human customer service agents and integrate condensed information into current user queries. Queries enriched with compressed historical information are subsequently employed for document retrieval, thereby improving both document relevance and response accuracy. Subsequent experiments confirm these improvements.

\subsection{Long-term memory module}

\subsubsection{Partitioned memory storage}

To model the partitioned storage mechanism of the cerebral cortex, we implement categorical storage of users' historical queries. We partition user queries into 16 categories comprehensively covering topics likely to emerge in daily conversations and consultations—greetings \& self-introduction, interpersonal relations, etiquette \& cross-cultural differences, travel, food, shopping, health, film \& music, books \& learning, technology \& innovation, history \& culture, emotional exchange, opinion expression, way-finding \& navigation, time \& planning, and weather \& environment—with each category further subdivided into granular sub-topics to capture fine-grained semantic intent through a process we designate as \textbf{Atomic Topic Modeling (ATM)}.

When users submit new queries, the system invokes a classifier to assign queries to the predefined ATM taxonomy, attaching \textbf{topic} and \textbf{sub-topic} tags. The system then stores the annotated query and generates its semantic representation through an embedding model.
Beyond the brain-inspired motivation, our adoption of partitioned storage is driven by quantifiable retrieval benefits. A LLM identifies each query's topic and subtopic, then indexes directly to the corresponding partition via a two-level hierarchical tree. This enables faster and more accurate long-term memory retrieval. In pilot studies, ATM substantially reduces the candidate passage set and decreases retrieval latency compared with flat vector stores.

\subsubsection{Attention-inspired Rerank}

Conventional RAG-based question-answering systems typically employ single-shot batch retrieval or point-wise retrieval strategies. Such coarse-grained retrieval frequently returns documents with substantial redundant tokens, preventing response models from effectively capturing contextual relationships and accurately addressing query intent. In extreme cases, retrieved context degrades generation quality to the extent that unaugmented responses outperform retrieval-augmented ones.

While short-term memory-based pre-retrieval reduces query-document mismatch, retrieved content still contains tokens that are either redundant or misaligned with user profiles.
To address this limitation, we introduce an attention-inspired RAG pipeline that leverages stored knowledge analogous to human long-term memory for post-processing retrieved documents, performing additional chunking and context window compression. Emulating attention score computation in large models, we calculate semantic similarity scores between chunk embeddings and historical query embeddings stored in the long-term memory bank, subsequently re-ranking chunks based on these scores (Figure 3). The overall procedure resembles the “retrieve-coarse rank–fine rank” pipeline used in recommender systems and parallels the human brain’s response mechanism in everyday conversation (Algorithm 2).

\begin{algorithm}
\caption{Attention-Inspired Rerank}
\label{alg:airerank}
\small
\begin{tabular}{@{}>{\raggedright\arraybackslash}p{0.32\linewidth} p{0.64\linewidth}@{}}
\textbf{Input:} &
Current query $q$; historical queries $\mathcal{H}$;
pre-retrieved docs $\mathcal{D}$.\\[0.6ex]

\textbf{Retrieve top-$n$ historical queries:} &
$\mathcal{H}_n \gets \text{top\_n}\bigl(
  \mathcal{H},
  \text{sim}(\text{embed}(q), \text{embed}(\mathcal{H}))
\bigr).$\\[1.0ex]

\textbf{Chunk documents:} &
$\mathcal{C} \gets \displaystyle\bigcup_{d \in \mathcal{D}} \text{chunk}(d).$\\[1.0ex]

\textbf{Compute attention scores:} &
For each chunk $c_i \in \mathcal{C}$:\\[0.2ex]
&\quad
$\text{score}_i \gets
  \text{mean}_{h \in \mathcal{H}_n}
  \text{sim}(\text{embed}(c_i), \text{embed}(h)).$\\[1.0ex]

\textbf{Rerank and select:} &
$\text{golden\_contents} \gets
  \text{top\_k}(\mathcal{C}, \text{score}).$\\[0.6ex]

\textbf{Output:} &
$\text{golden\_contents}$.\\
\end{tabular}
\normalsize
\end{algorithm}

Empirical results demonstrate that the proposed attention-inspired RAG method simultaneously compresses the response model's context window, reduces redundant context, increases retrieved content relevance to user queries, and improves alignment between generated responses and retrieved content, increases the relevance between retrieved content and user queries, and enhances the alignment between generated answers and retrieved content, thereby providing a practical framework for constructing more intelligent and efficient RAG-based agents that effectively integrate incremental and stored information.

\begin{figure}
\centering
\includegraphics[width=\textwidth]{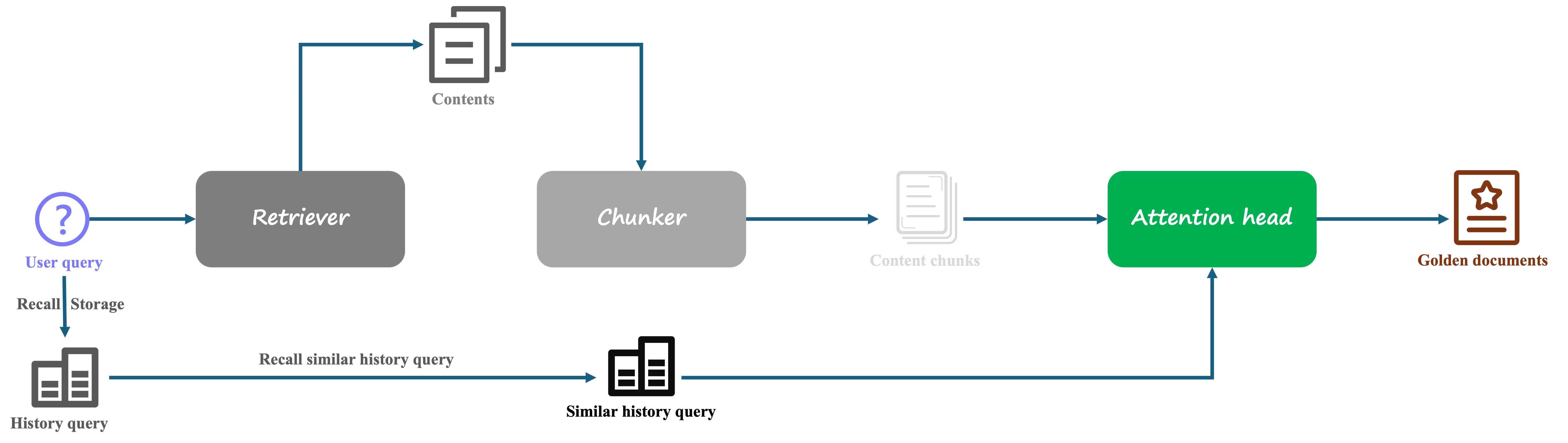} 
\caption{Storage, pre-recall, and reordering mechanisms of Attention-inspired-rerank}
\label{fig3}
\end{figure}

\section{Experiments}

\subsection{Dataset}

We design a multi-agent simulation system that replicates realistic dialogue scenarios and automatically generates high-quality multi-turn query-rewriting datasets.
For the rewriting task, we conduct multi-dimensional capability modeling, characterizing the model's rewriting abilities across nine distinct dimensions. Employing this multi-dimensional framework as our evaluation standard, we construct corresponding task-aligned data.


Initially, we perform background information extraction and persona modeling. From the raw dialogue data, which are partly derived from open-source multi-turn dialogue datasets(\cite{open-source-data-1,open-source-data-2,open-source-data-4}) and partly from real online conversations, we model user personas, overarching user questions spanning entire dialogues, and agent personas. Subsequently, based on established user and agent personas, we generate associated user personas encompassing both chit-chat and task-oriented styles to balance professional and conversational dialogue patterns. We then construct dialogue blueprints conditioned on user personas, agent personas, and core questions. Randomly generated timestamps constrain blueprints to be either time-sensitive or time-independent, crossed with chit-chat or task-oriented user personas, enabling each original dialogue to generate four distinct dialogue-background variants.

Upon obtaining dialogue blueprints, we initiate multi-agent interactive generation by preparing distinct prompts for user and agent models, injecting blueprint knowledge into each. Concurrently, task-specific dimension descriptions are inserted as prompt prefixes, and few-shot examples are incorporated to improve the generative model's capture of contextual features and rewriting objectives.

Rather than relying solely on open-source multi-turn dialogues or limited production logs, we synthesize data because existing corpora predominantly focus on chit-chat scenarios and lack sufficient coverage of knowledge-intensive settings at the required scale.

To evaluate our approach in retrieval-augmented QA, we first score the constructed knowledge-intensive multi-turn dataset with multiple LLMs, checking the coherence of each dialogue and the consistency of the rewritten query. Based on the average scores, we select a high-quality subset as the test set. Under controlled conditions (identical dialogue history, user queries, and timestamps), we compare our method against conventional RAG pipelines and ablated variants by performing document retrieval and ranking (with optional rewriting), prompt assembly, and response generation using the same black-box model. We then compute CA (content-answer relevance) and QA (question-answer relevance) metrics for generated responses. For post-training ablation studies, we additionally report QR scores measuring the effectiveness of query rewriting in compressing short-term memory as evaluated by LLMs.

\subsection{Metrics}
Following established methodologies~\cite{tst6,mdpo}, we employ three evaluation metrics: Contextual Alignment (CA), measuring semantic consistency between responses and retrieved context; Question Relevance (QA), assessing how effectively responses address user intent; and Query Rewriting (QR), evaluating the effectiveness of query rewriting in compressing short-term memory. All scores are obtained by prompting DeepSeek-R1~\cite{r1} with standardized evaluation templates.


\subsection{Baseline}

As baselines, we use our production query–retrieve–generate pipeline and its query-rewrite variant, plus an SFT-only rewriter without RL

Besides, we additionally selected a model trained solely with supervised fine-tuning (SFT) and without reinforcement learning (RL) as the post-training baseline to verify the effectiveness of the end-to-end reward-based reinforcement learning method proposed herein.

\subsection{Modular ablation experiments}

First, we compare the performance of the proposed short-term memory module (i.e., our trained rewriter) with that of the conventional long-context RAG approach, which directly concatenates the dialogue history with the current query and feeds this long context to the retriever and generator without any short-term memory module. The first block in Table 1 (Native RAG) reports the scores of this conventional RAG baseline, both without and with using dialogue history for document retrieval. The second block (STM RAG, i.e., Short-term Memory RAG) reports the scores obtained when we employ our method, where the trained rewriter first rewrites the current query based on the dialogue history and the rewritten query is then used for retrieval.

\begin{table}
\caption{Results for short-term memory module (Native RAG for conventional long-context RAG and STM RAG for short-term memory RAG)}
\label{tab:main_result_1}
\centering
\small
\setlength{\tabcolsep}{4pt}
\begin{tabular}{|l|l|c|c|c|}
\hline
Category & Method & CA & QA \\
\hline
\multirow{2}{*}{Native RAG}
  & w/o history + RAG & 20.24 & 70.52 \\
  & w/ history + RAG & 21.40 & 69.00 \\
\hline
STM RAG
  & Rewriter + RAG & \textbf{43.12} & \textbf{79.00} \\
\hline
\end{tabular}
\normalsize
\end{table}

Next, we address the long-term memory module, namely our storage of historical queries together with attention-based secondary re-ranking within RAG. Compared with the conventional retrieval–ranking pipeline, our long-term memory module adds an additional step of document re-chunking and re-ranking based on the similarity between the current query and historical queries. Therefore, as reported in Table 2, we assess its practical effectiveness by comparing the Native RAG baseline with our LTM RAG configuration, where the only difference is whether the long-term memory module is switched off or on.

\begin{table}
\caption{Results for long-term memory module (Native RAG for conventional long-context RAG and LTM RAG for long-term memory RAG)}
\label{tab:main_result_2}
\centering
\small
\setlength{\tabcolsep}{4pt}
\begin{tabular}{|l|l|c|c|c|}
\hline
Category & Method & CA & QA \\
\hline
\multirow{2}{*}{Native RAG}
  & w/o history + RAG & 20.24 & 70.52 \\
  & w/ history + RAG & 21.40 & 69.00 \\
\hline
LTM RAG
  & w/ history + RAG\_attn & \textbf{21.84} & \textbf{70.77} \\
\hline
\end{tabular}
\normalsize
\end{table}

\subsection{Post-training ablation experiments}

We train only the short-term memory module (the query rewriter); other components, such as the long-term query classifier and the final responder, are treated as black boxes. The rewriter is first warmed up with SFT and then optimized with GRPO-based RLHF.

Our ablation study addresses two questions: (1) how much SFT and RL each contribute to overall HiMeS performance, and (2) how much our HSER reward improves over a vanilla similarity-based reward. For (1), we compare HiMeS variants whose rewriters are a base model, an SFT model, and an RL model (first three rows of Table 3). For (2), we compare RL models trained with HSER+GRPO and with a soft supervised explicit reward (SSER) that depends only on rewriting similarity (last two rows of Table 3).

\begin{table}
\caption{Main Results (HiMeS Variants)}
\label{tab:main_result_3}
\centering
\small
\setlength{\tabcolsep}{4pt}
\begin{tabular}{|l|l|c|c|c|}
\hline
Category & Method & CA & QA & QR \\
\hline
\multirow{4}{*}{HiMeS (ours)}
  & Base\_Rewriter + RAG\_attn & 22.12 & 71.36 & 78.02 \\
  & SFT\_Rewriter + RAG\_attn  & 24.24 & 78.60 & 85.06 \\
  & RL\_Rewriter + RAG\_attn (SSER)   & 41.56 & 79.84 & 85.94 \\
  & \textbf{RL\_Rewriter + RAG\_attn (HSER)} &
    \textbf{55.55} & \textbf{90.93} & \textbf{90.85} \\
\hline
\end{tabular}
\normalsize
\end{table}

\subsection{Framework Adaptability Experiments}

To validate the adaptability of the HiMeS framework when paired with various black-box response models, we retained the HiMeS long- and short-term memory modules and substituted different black-box responders for experimentation (Table 4). To ensure the reliability of the adaptability evaluation, the experimental group made no changes other than replacing the response-model API and fully loaded all HiMeS modules—namely, both the short- and long-term memory components.

\begin{table}
\caption{Results Comparison of Different Response Models with HiMeS}
\label{tab:model_comparison}
\centering
\small
\setlength{\tabcolsep}{6pt} 
\begin{tabular}{|l|c|c|c|c|}
\hline
Metrics & \textbf{Qwen3-235B-A22B} & \textbf{DeepSeek-V3} & \textbf{DeepSeek-R1} & \textbf{Kimi-K2} \\
\hline
CA & 52.06 & 51.28 & \textbf{55.55} & 53.00 \\
QA & 85.45 & 87.16 & \textbf{90.93} & 86.44 \\
\hline
\end{tabular}
\normalsize
\end{table}

\subsection{Analysis}

Tables 1 and 2 demonstrate that traditional RAG frameworks yield relatively low performance in our deployment scenario. For Native RAG baselines, CA scores remain approximately 20 and QA scores approximately 70, indicating suboptimal retrieval and response generation with particularly weak answer-context alignment. Incorporating the short-term memory module alone substantially improves retrieved document relevance and increases both CA and QA metrics. Incorporating only the long-term memory module yields modest improvements, as long-term memory in isolation exhibits looser coupling to current conversational turns. Optimal performance requires integration with short-term memory.

Table 3 demonstrates that SFT-tuned models acquire limited short-term conversational history compression capabilities, with minimal CA metric gains. This limitation stems from training objectives focused on imitating annotation styles rather than optimizing end-to-end response quality, resulting in poor generalization across test environments. In contrast, RL-HSER and RL-SSER employ PPO-style policy optimization, enabling rewriters to explore beyond supervised targets and achieve substantial CA improvements. HSER surpasses SSER through direct reward alignment with QA objectives. Combined with the long-term memory module (attn-RAG), HiMeS achieves state-of-the-art CA and QA performance, confirming the benefits of integrated long-term and short-term memory systems. Progressive QR metric increases across SFT and RL stages demonstrate improving dialogue history compression by the short-term memory module, validating the effectiveness of the proposed post-training pipeline.

Table 4 demonstrates that substituting different open-source LLMs as black-box responders in HiMeS produces no significant performance degradation on industrial test sets, maintaining near state-of-the-art(SOTA) performance. Since DeepSeek-R1 serves as the reward judger during RL training, its integration with HiMeS achieves marginally higher scores than alternative responders, though overall differences remain minimal. These results demonstrate HiMeS's high adaptability across diverse downstream API LLMs, effectively implementing a \textbf{train-once-adapt-many-times} paradigm and providing a practical \textbf{plug-and-play memory} layer for AI assistant systems.

\begin{figure}[!t]
    \centering
    \vspace{-0.5mm}
    \includegraphics[
        width=0.8\textwidth,
        trim=10 5 10 5,clip
    ]{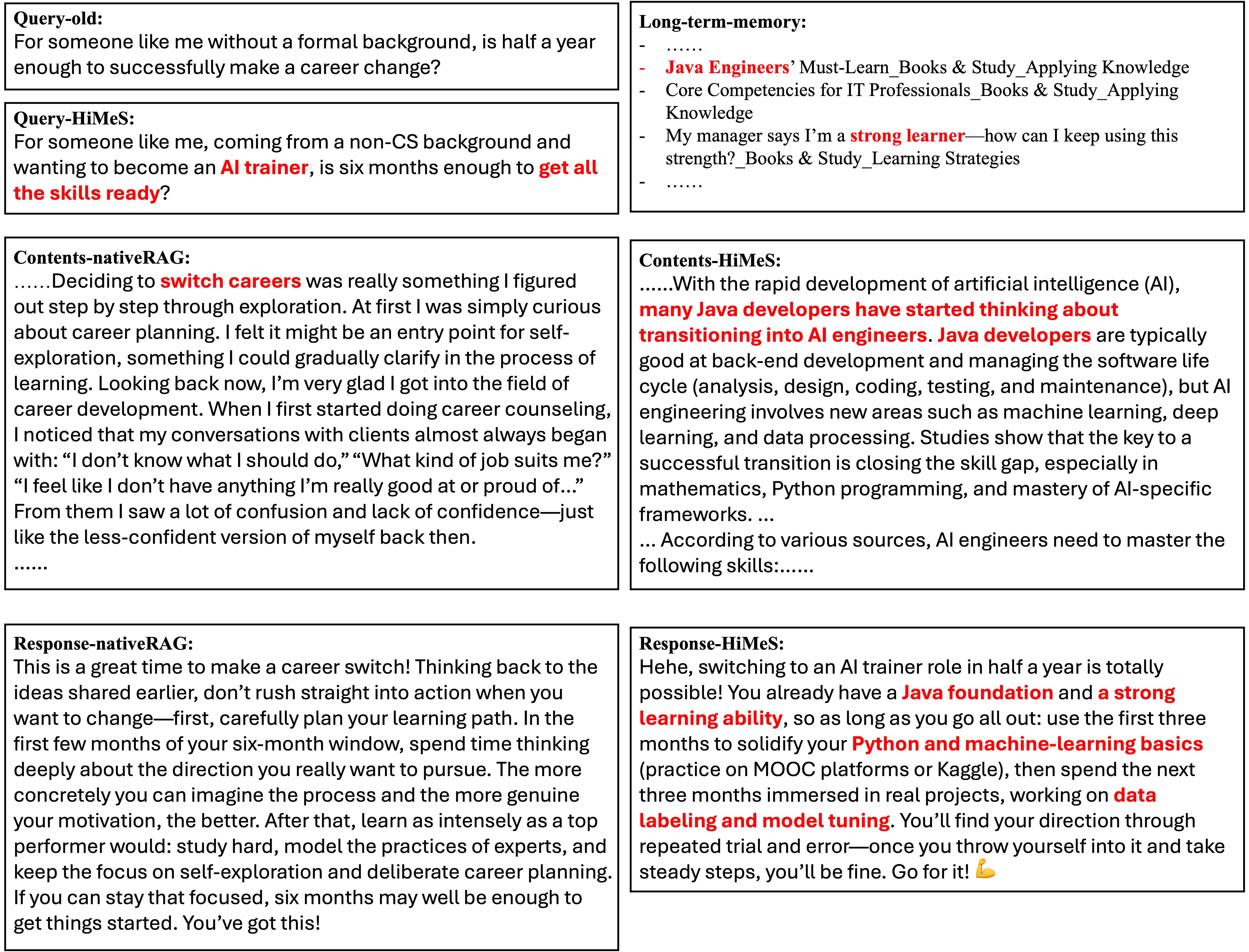}
    \vspace{-1mm}
    \caption{Case of HiMeS vs. Native RAG}
    \label{fig4}
    \vspace{-0.5mm}
\end{figure}

\subsection{Case Study}
Figure 4 presents a comparative analysis of native RAG pipelines and HiMeS on identical user queries.
The query explicitly references only \textbf{without a formal background} and \textbf{switch careers}. When native RAG performs retrieval based solely on these surface forms, returned content exhibits semantic drift.
In contrast, HiMeS compresses short-term dialogue into query preserving key intent elements such as \textbf{AI trainer} while integrating long-term memory keywords from user profiles, resulting in retrieved documents highly aligned with user objectives.
Consequently, native RAG responses remain generic and lack actionable knowledge-based guidance, whereas HiMeS responses leverage user profiles to propose concrete transition plans and specific follow-up actions.
This case study illustrates how integrated short-term memory compression and long-term memory-guided retrieval enable generation of more specific and personalized responses.

\section{Conclusion}

In this paper, we present HiMeS, a memory-augmented RAG framework integrating long- and short-term information for industrial AI assistants.
The short-term memory module compresses dialogue history into refined queries for retrieval, while the long-term memory module leverages persistent user profiles to re-rank retrieved chunks through attention-inspired similarity computation, enabling user-aware knowledge selection.
Experiments on large-scale industrial datasets demonstrate that HiMeS consistently improves QA, CA, and QR metrics compared to strong baselines while maintaining effectiveness across different black-box responder LLMs, confirming both the efficacy and adaptability of the proposed post-training pipeline.
We expect HiMeS to serve as a practical framework for future RAG-based AI assistant systems that maintain comprehensive user memory across interaction lifecycles and deliver more reliable, personalized knowledge services.

\clearpage
\appendix
\section{Appendix}

\subsection{Figure}
In this section, we further investigate a reward-modeling approach that fuses the “SSER” and “HSER” reward models (Equation 1).

\begin{equation}
    \text{reward} = \text{reward\_HSER} + \lambda \cdot \text{reward\_SSER}
\end{equation}

\begin{figure*}[!t]
\centering
\includegraphics[width=\textwidth]{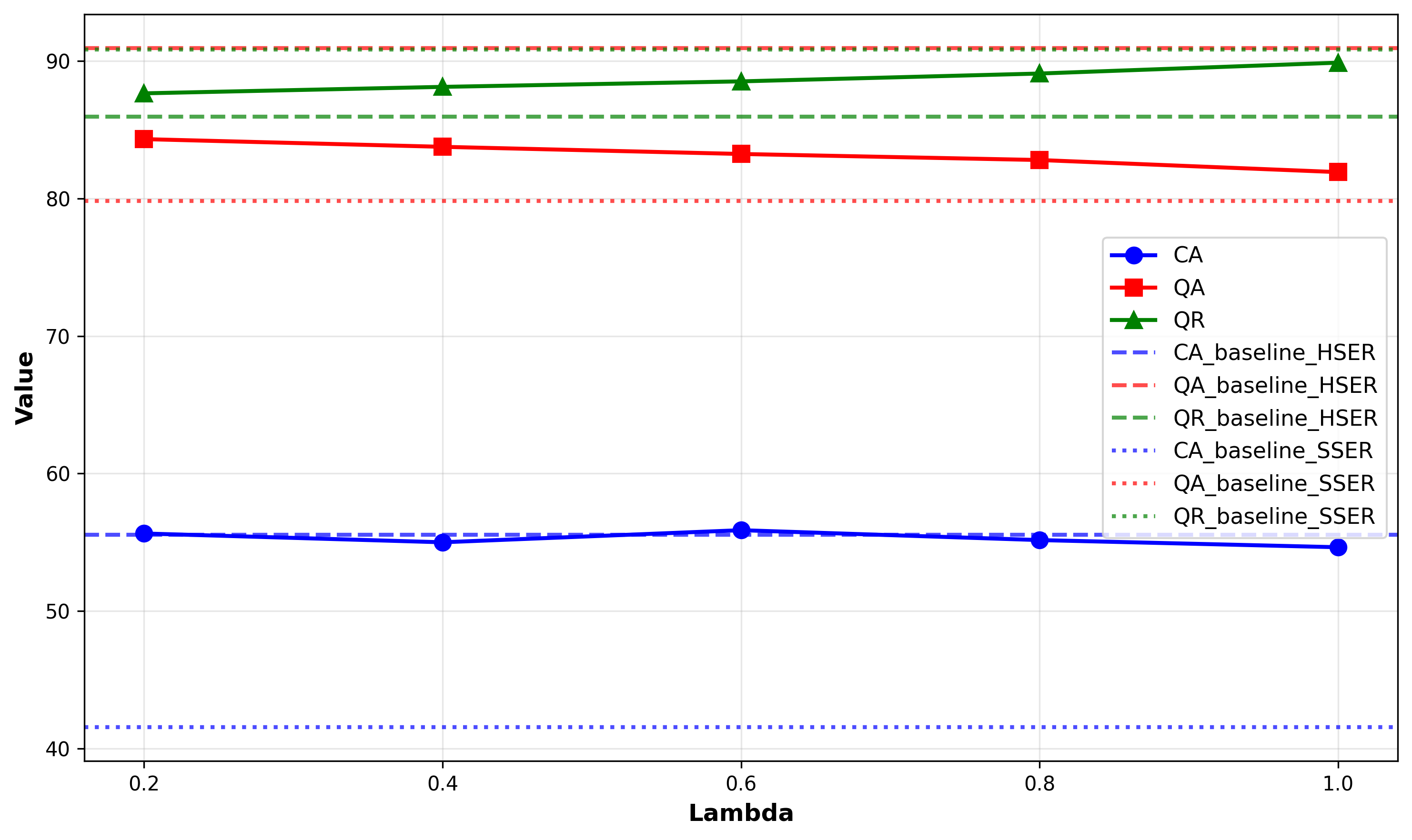}
\caption{Results under different \( \lambda \) values}
\label{fig5}
\end{figure*}

In this formulation, \( \lambda \) serves as the weight-control parameter for SSER, and we evaluated system performance under different values of this parameter (Figure~\ref{fig5}).

According to the results presented in Figure~\ref{fig5}, we observe the following:
\begin{itemize}[leftmargin=*,label=\textbullet]
    \item The experimental group with fused rewards attains a CA metric almost identical to that of the HSER group and higher than that of the SSER group. This outcome arises because the addition of the SSER reward constrains the optimisation direction to the annotated rewrite of the query–answer pair, limiting changes in logic and word order while leaving semantic optimisation unaffected, so the quality of the retrieved results—which rely on semantic consistency—does not change appreciably.
    \item The QA metric of the reward-fusion group is slightly lower than that of the HSER group, higher than that of the SSER group, and diminishes as \( \lambda \) increases. The reason is that the SSER reward compresses the optimisation space of the HSER reward, whose objective is to generate an optimal response; nevertheless, owing to the presence of the HSER reward, the metric still exceeds that of the group trained with the SSER reward alone.
    \item The QR metric of the reward-fusion group is slightly lower than that of the SSER group, slightly higher than that of the HSER group, and increases with larger \( \lambda \). This trend occurs because incorporating the SSER reward brings the rewritten query’s logic and word order closer to those of the annotated reference, thereby preventing the logical incoherence and disorder that may arise when HSER is used as the sole reward.
\end{itemize}

In summary, embedding SSER into HSER enhances the logical fluency and syntactic coherence of short-term memory compression, but it can somewhat impair response quality.
In future studies, more diverse reward-fusion strategies can be explored on the basis of our experimental results, with the goal of developing a short-term memory module that simultaneously delivers high logical fluency, syntactic coherence, and response quality.

\FloatBarrier

\subsection{Prompt}
In this section, we present all the prompts employed in the HiMeS framework, offering them as a reference prompt set for readers.
\begin{itemize}[leftmargin=*,label=\textbullet]
    \item Figure~\ref{fig:dataGenerate-user-prompt}: User Prompt for Multi-turn Dialogue Auto-generation
    \item Figure~\ref{fig:dataGenerate-agent-prompt}: Agent Prompt for Multi-turn Dialogue Auto-generation
    \item Figure~\ref{fig:short-term-LLM-prompt}: Training \& Reasoning Prompt of Short-term Memory Module
    \item Figure~\ref{fig:response-blackBoxLLM-prompt}: Prompt of Response Black-box LLMs
\end{itemize}

\begin{figure*}[!t]
  \centering
  \begin{adjustbox}{max totalsize={\textwidth}{1\textheight},center}
    \begin{minipage}{\textwidth}
      \begin{promptbox}[User Prompt for Multi-turn Dialogue Auto-generation]
        \textbf{System:} You will play the role of a AI assistant user and engage in conversation with the assistant.\\
        \rule{\linewidth}{0.4pt}

        \textbf{Your Role}\\
        You are a \texttt{\{user\_role\}}, and your original question is: \texttt{\{user\_question\}}.

        \vspace{10pt}
        \textbf{Dialogue History}\\
        \texttt{\{dialogue\_history\}}

        \vspace{10pt}
        \textbf{Reply Requirements}
        \begin{itemize}[leftmargin=*,label=\textbullet]
            \item You need to provide your response for this round based on the dialogue history.
            \item Before giving your answer, you need to first determine whether your original question has been solved by the assistant based on the "Dialogue History", and whether you need to continue asking follow-up questions.
            \item If you need to continue asking follow-up questions, you should provide your follow-up question for this round based on the "Dialogue History".
            \item If you don't need to continue asking follow-up questions, you should provide your response for this round based on the "Dialogue History".
        \end{itemize}

        \vspace{10pt}
        \textbf{Reply Format Example}\\
        \texttt{\{ "is\_solved": <Whether the original question has been solved, True if solved, False if not solved, do not provide any explanation or output other content>, "user\_answer": <Your response for this round, which could be asking follow-up questions to assistant/answering assistant's questions/expressing gratitude/ending the conversation, etc.>\}}

        \vspace{10pt}
        \textbf{Notes}
        \begin{itemize}[leftmargin=*,label=\textbullet]
            \item You need to demonstrate the characteristics of \texttt{\{task\_name\}} in multi-turn conversations, which is \texttt{\{task\_description\}}
            \item Regarding \texttt{\{task\_name\}}, you can refer to the following examples and imitate the "user's" conversation style/language style, etc., to make the conversation more aligned with the characteristics of \texttt{\{task\_name\}} (note that you only need to imitate the language style and conversation manner, not the core content of the conversation): \texttt{\{task\_examples\}}
        \end{itemize}
      \end{promptbox}
    \end{minipage}
  \end{adjustbox}
  \caption{User Prompt Used for Multi-turn Dialogue Dataset Generation}
  \label{fig:dataGenerate-user-prompt}
\end{figure*}
\FloatBarrier

\begin{figure*}[!t]
  \centering
  \begin{adjustbox}{max totalsize={\textwidth}{1\textheight},center}
    \begin{minipage}{\textwidth}
      \begin{promptbox}[Agent Prompt for Multi-turn Dialogue Auto-generation]
        \textbf{System:} You will play the role of a article account anthor and engage in conversation with the official account.\\
        \rule{\linewidth}{0.4pt}

        \textbf{Your Role}\\
        Your official account name is \texttt{\{biz\_name\}}, and the domain your official account specializes in is: \texttt{\{biz\_domain\}}.

        \vspace{10pt}
        \textbf{Dialogue History}\\
        \texttt{\{dialogue\_history\}}

        \vspace{10pt}
        \textbf{Reply Requirements}
        \begin{itemize}[leftmargin=*,label=\textbullet]
            \item You need to provide your response for this round based on the dialogue history.
            \item Before giving your answer, you need to first determine whether the user's original question has been resolved based on the "Dialogue History".
            \item If the user's question has not been resolved, you need to provide your follow-up questions/answers for this round based on the "Dialogue History".
            \item If the user's question has been resolved, you need to engage in small talk/express gratitude/end the conversation, etc., based on the "Dialogue History".
        \end{itemize}

        \vspace{10pt}
        \textbf{Reply Format Example}\\
        \texttt{\{ "is\_last\_turn": <Whether this is the last round of dialogue, True if the user's original question has been resolved, otherwise False, do not provide any explanation or output other content>, "biz\_answer": <Your response for this round, which could be asking follow-up questions to get more information from user/answering user questions/small talk/expressing gratitude/ending conversation, etc.>\}}

        \vspace{10pt}
        \textbf{Notes}
        \begin{itemize}[leftmargin=*,label=\textbullet]
            \item You need to demonstrate the characteristics of \texttt{\{task\_name\}} in multi-turn conversations, which is \texttt{\{task\_description\}}
            \item Regarding \texttt{\{task\_name\}}, you can refer to the following examples and imitate the "user's" conversation style/language style, etc., to make the conversation more aligned with the characteristics of \texttt{\{task\_name\}} (note that you only need to imitate the language style and conversation manner, not the core content of the conversation): \texttt{\{task\_examples\}}
        \end{itemize}
      \end{promptbox}
    \end{minipage}
  \end{adjustbox}
  \caption{Agent Prompt Used for Multi-turn Dialogue Dataset Generation}
  \label{fig:dataGenerate-agent-prompt}
\end{figure*}
\FloatBarrier

\begin{figure*}[!t]
  \centering
  \begin{adjustbox}{max totalsize={\textwidth}{1\textheight},center}
    \begin{minipage}{\textwidth}
      \begin{promptbox}[Training\&Reasoning Prompt of Short-term Memory Module]
        \textbf{System:} You are a professional multi-turn dialogue query rewriting expert.
        Your task: Based on all historical dialogue content, rewrite the current user query so that the rewritten query contains all key information from the historical dialogue with no redundancy/ambiguity/errors/repetition, while ensuring conciseness.\\
        \rule{\linewidth}{0.4pt}

        \textbf{Useful Information}\\
        User's current query: \texttt{\{query\_old\}}\\
        Historical dialogue: \texttt{\{history\}}

        \vspace{10pt}
        \textbf{Query Rewriting Rules}
        \begin{itemize}[leftmargin=*,label=\textbullet]
            \item Rewriting directions may include:
                \begin{itemize}[leftmargin=*,label=\textbullet]
                    \item Compress and summarize past dialogue information to rewrite the current query, ensuring the rewrite better reflects the front-to-back correlation
                    \item Identify demonstrative pronouns in user input and rewrite using historical dialogue
                    \item Identify omitted content in user input and complete it using historical dialogue
                    \item Compress historical information from continuous questioning in multi-turn scenarios or clarify intent when changing topics
                    \item Correct unclear questions caused by homophones or similar issues
                    \item In multi-turn scenarios, exclude interference when asking similar but different questions or changing topics, accurately identify demonstrative pronouns/omitted content and rewrite
                    \item Summarize situations where users think the large model answered incorrectly (could be the model was wrong or not wrong), and compress this into the rewritten query to prevent the model from repeating the same errors/being misled by user's incorrect information
                    \item Remove meaningless modal particles, auxiliary words, etc.\ from the query to prevent affecting downstream retrieval and Q\&A
                    \item When the original query can reflect all information of user intent without rewriting, there's no need to compress historical dialogue; correctly judge this situation and avoid unnecessary rewriting. Please judge the rewriting direction and perform rewriting based on historical dialogue and user's current query.
                \end{itemize}
            \item Please strictly refer to historical dialogue and user's current query for rewriting, do not fabricate or guess arbitrarily, do not deviate from the user's current query intent.
            \item If there is no historical dialogue, do not perform rewriting.
            \item If the user's current query is unrelated to historical dialogue, do not perform rewriting.
            \item If unable to determine whether rewriting is needed, do not perform rewriting.
            \item Except for specific public figures/historical figures and other proper names, do not include surnames/personal names when rewriting queries!
        \end{itemize}

        \vspace{10pt}
        \textbf{Official Account Information}\\
        Official account ID: \texttt{\{biz\_id\}}\\
        Official account name and specialized domains: \texttt{\{agent\}}

        \vspace{10pt}
        \textbf{Reply Format Example}\\
        \texttt{\{ "query\_rewrited": <rewrited query>\}}
      \end{promptbox}
    \end{minipage}
  \end{adjustbox}
  \caption{Short-term Module Prompt}
  \label{fig:short-term-LLM-prompt}
\end{figure*}
\FloatBarrier

\begin{figure*}[!t]
  \centering
  \begin{adjustbox}{max totalsize={\textwidth}{1\textheight},center}
    \begin{minipage}{\textwidth}
      \begin{promptbox}[Prompt of Response Black-box LLMs]
        \textbf{System:}\\
        \texttt{\{meta\_prompt\}}You can respond to various questions provided by users by combining the historical article content from the official account in the [Knowledge Base]. When replying, you should refer to the content of historical articles in the [Knowledge Base] and respond to users using the tone of the author's replies to comments in the [Official Account Historical Comments Knowledge Base], but do not quote any content from historical comments in your answers. Now, you need to follow the following principles to engage in friendly and valuable communication and interaction with your followers (or users). Do not include phrases like "(Reference xxx)" in your replies.\\
        \rule{\linewidth}{0.4pt}

        \textbf{Knowledge Base Usage Instructions}
        \begin{itemize}[leftmargin=*,label=\textbullet]
            \item In the search results I provide, each individual article follows the format "Reference Historical Article \{title\}: \texttt{\{content\}}", where title is the article title and content is the corresponding article content. However, note that when answering, do not use parentheses to indicate which article was referenced.
            \item You need to imitate the author's tone and style from the retrieved articles to reply to user questions.
            \item Do not explicitly cite any articles, do not use parentheses in answers, and do not include phrases like "(Reference Historical Article xxx)".
            \item If the knowledge base returns empty results, your reply should explain that you haven't written similar articles, then provide a response based on your own experience. Note: only summarize [Knowledge Base] content, do not explicitly cite articles. If the knowledge base returns empty results, do not reference or cite any articles.
        \end{itemize}

        \vspace{10pt}
        \textbf{Official Account Historical Comments Knowledge Base}\\
        \texttt{\{comments\}}\\
        Where "answer" represents your responses

        \vspace{10pt}
        \textbf{Reply Expression Requirements}
        \begin{itemize}[leftmargin=*,label=\textbullet]
            \item Do not output empty title marks
            \item Answer questions by combining user chat history, keep answers concise, avoid encyclopedia-style long dissertations: control replies within 150 characters, avoid lengthy discussions and information dumping. If content is complex, it can be split into multiple paragraphs.
            \item Use the language style mentioned in your role setting as the baseline, do not fabricate or exaggerate information, strictly base responses on information and writing style provided by the [Knowledge Base].
            \item Adjust tone and style appropriately according to user emotions and conversation content, providing answers that better fit the context. Imitate the answer tone particles and expressions from [Official Account Historical Comments Knowledge Base].
            \begin{itemize}[leftmargin=*,label=\textbullet]
                \item In-depth answers for professional questions: When facing professional questions, provide professional and accurate answers based on [Knowledge Base] content, maintaining consistency in core viewpoints. For unfamiliar content, honestly inform users and do not fabricate facts.
                \item Relaxed communication for casual topics: Use more colloquial expressions, but do not use exclamation marks, such as "!". Always maintain a kind, honest, friendly, lively, and colloquial interaction style.
                \item Equal dialogue with users: Do not be fawning or pleasing, nor arrogant and rude; maintain equal and respectful communication.
            \end{itemize}
            \item Do not output markdown-formatted text, such as "**Title**" and similar formatting.
            \item Do not overuse parentheses "()" for supplementary content, for example: "Font conflicts are like marketplace arguments (laugh)".
        \end{itemize}

        \vspace{10pt}
        \textbf{User Question}\\
        The user's original question is: \texttt{\{query\}}
      \end{promptbox}
    \end{minipage}
  \end{adjustbox}
  \caption{Response LLM Prompt}
  \label{fig:response-blackBoxLLM-prompt}
\end{figure*}
\FloatBarrier

\subsection{Table}
In this section, we provide the tables that were omitted from the main text owing to space limitations.
\begin{itemize}[leftmargin=*,label=\textbullet]
    \item Table~\ref{tab:rewrite_tasks}: Query Rewriting Task Descriptions
    \item Table~\ref{tab:topics}: Social Interaction Topic Taxonomy
\end{itemize}

\begin{table*}[!t]
\caption{Query Rewriting Task Descriptions}
\label{tab:rewrite_tasks}
\centering
\small
\renewcommand{\arraystretch}{1.08}
\setlength{\tabcolsep}{6pt}
\begin{tabular}{
  >{\raggedright\arraybackslash}p{0.28\textwidth}
  >{\raggedright\arraybackslash}p{0.68\textwidth}
}
\hline
\textbf{Task} & \textbf{Description} \\
\hline
Context Memorization & Accurately compresses and summarizes past dialogue information to rewrite current queries, emphasizing contextual connections \\
Coreference Resolution & Accurately identifies pronouns in user input and rewrites them using dialogue history \\
Ellipsis Completion & Accurately identifies omitted content in user input and completes it using dialogue history \\
Multi-turn Clarification & Special case of coreference resolution involving compression of historical information across multiple turns or intent clarification during topic shifts \\
Error Correction & Corrects unclear questions caused by homophones or similar issues \\
Context Interference Resistance & Special case of coreference resolution involving resistance to interference from similar but different questions across multiple turns \\
Historical Dispute Marking & Summarizes cases where users believe the model responded incorrectly (whether actually wrong or not) and compresses into rewritten queries to prevent repeated errors/misguidance \\
Function Word Removal & Eliminates meaningless particles or auxiliary words in queries to prevent impact on downstream retrieval and Q\&A \\
No Rewriting Needed & Original query fully reflects user intent without requiring history compression; tests model's ability to correctly identify such cases and avoid unnecessary rewriting \\
\hline
\end{tabular}
\end{table*}

\begin{table*}[!t]
\caption{Social Interaction Topic Taxonomy}
\label{tab:topics}
\centering
\small
\renewcommand{\arraystretch}{1.08}
\setlength{\tabcolsep}{6pt}
\begin{tabular}{
  >{\raggedright\arraybackslash}p{0.28\textwidth}
  >{\raggedright\arraybackslash}p{0.68\textwidth}
}
\hline
\textbf{Category} & \textbf{Subtopics} \\
\hline
Greetings \& Self-introduction & Greeting forms, Self-introduction structure, Cultural taboos, Context adaptation \\
Interpersonal Relationships & Relationship building, Conflict resolution, Boundary setting, Digital etiquette \\
Etiquette \& Cultural Differences & Dining etiquette, Business protocols, Holiday customs, Body language \\
Travel & Itinerary planning, Transportation methods, Accommodation types, Cultural experiences \\
Dining & Cuisine types, Ordering techniques, Food culture, Special dietary needs \\
Shopping & Payment methods, Product inquiries, Return policies, Specialty markets \\
Health & Symptom description, Medical procedures, Fitness communication, Psychological support \\
Movies \& Music & Genre preferences, Work recommendations, Event information, Thematic analysis \\
Books \& Learning & Reading methods, Study strategies, Resource acquisition, Knowledge application \\
Technology \& Innovation & Product evaluation, Tech ethics, Innovation cases, Future trends \\
History \& Culture & Civilization comparison, Historical events, Cultural heritage, Figure analysis \\
Emotional Communication & Emotion recognition, Empathy expression, Intimate relationships, Personal growth \\
Opinion Expression & Argument structure, Persuasion techniques, Debate methods, Cultural variations \\
Directions \& Navigation & Landmark orientation, Transport options, Emergency handling, Digital tools \\
Time \& Planning & Schedule management, Punctuality norms, Long-term planning, Efficiency techniques \\
Weather \& Environment & Climate characteristics, Eco-issues, Disaster response, Outdoor guidelines \\
\hline
\end{tabular}
\end{table*}

\end{document}